\documentclass[11pt,a4paper]{article}

\usepackage[margin=1in]{geometry}
\usepackage[T1]{fontenc}
\usepackage[utf8]{inputenc}
\usepackage{lmodern}
\usepackage{authblk}

\usepackage{amsmath,amssymb,amsthm}

\usepackage{xcolor}
\usepackage{graphicx}
\usepackage{booktabs}
\usepackage{multirow}
\usepackage{array}
\usepackage{tabularx}
\usepackage{enumitem}
\usepackage{float}
\usepackage{setspace}
\usepackage{parskip}

\usepackage{footmisc}

\usepackage{microtype}
\microtypesetup{nopatch=footnote}

\usepackage{caption}
\usepackage{subcaption}

\usepackage{tikz}
\usetikzlibrary{shapes.geometric,arrows.meta,positioning,fit,backgrounds,calc,patterns}

\usepackage{listings}
\usepackage{algorithm}
\usepackage{algpseudocode}

\usepackage{mdframed}

\usepackage{authblk}

\usepackage{hyperref}
\usepackage{cleveref}

\newtheorem{definition}{Definition}

\hypersetup{
  colorlinks=true,
  linkcolor=blue!55!black,
  citecolor=blue!55!black,
  urlcolor=blue!55!black,
}

\definecolor{codebg}{rgb}{0.96,0.96,0.96}
\definecolor{codegreen}{rgb}{0.0,0.42,0.0}
\definecolor{codepurple}{rgb}{0.42,0.0,0.52}
\definecolor{codeorange}{rgb}{0.7,0.28,0.0}
\definecolor{codered}{rgb}{0.7,0.0,0.0}

\lstdefinestyle{python}{
  backgroundcolor=\color{codebg},
  commentstyle=\color{codegreen}\itshape,
  keywordstyle=\color{codepurple}\bfseries,
  stringstyle=\color{codeorange},
  basicstyle=\ttfamily\small,
  breaklines=true,
  keepspaces=true,
  language=Python,
  showstringspaces=false,
  tabsize=4,
  frame=single,
  rulecolor=\color{black!20},
  xleftmargin=6pt, xrightmargin=6pt,
}
\lstdefinestyle{json}{
  backgroundcolor=\color{codebg},
  basicstyle=\ttfamily\small,
  breaklines=true,
  frame=single,
  rulecolor=\color{black!20},
  xleftmargin=6pt, xrightmargin=6pt,
  morestring=[b]",
  stringstyle=\color{codeorange},
}

\tikzset{
  task/.style={rectangle, rounded corners=4pt, draw=black!60,
               fill=blue!12, minimum width=2.5cm, minimum height=0.95cm,
               font=\small, text centered, text width=2.3cm},
  gateway/.style={diamond, draw=black!60, fill=orange!18, aspect=1.7,
                  minimum width=1.4cm, font=\scriptsize, text centered},
  hook/.style={rectangle, rounded corners=2pt, draw=red!65, fill=red!8,
               minimum width=2.2cm, minimum height=0.75cm,
               font=\scriptsize\itshape, text centered, text width=2.0cm,
               densely dashed, line width=0.8pt},
  event/.style={circle, draw=black!60, fill=white,
                minimum size=0.65cm, font=\scriptsize},
  flow/.style={-{Stealth[length=5.5pt]}, thick, black!65},
  hookflow/.style={-{Stealth[length=5.5pt]}, thick, red!60, densely dashed},
  policy/.style={rectangle, rounded corners=2pt, draw=green!50!black,
                 fill=green!6, minimum width=1.8cm, minimum height=0.6cm,
                 font=\scriptsize, text centered, text width=1.7cm},
  agent/.style={rectangle, rounded corners=4pt, draw=purple!55,
                fill=purple!8, minimum width=2.6cm, minimum height=1cm,
                font=\small, text centered, text width=2.4cm},
  frame/.style={rectangle, rounded corners=6pt, draw=teal!60,
                fill=teal!5, line width=1.2pt},
}

\newmdenv[linewidth=1pt,linecolor=blue!40,backgroundcolor=blue!3,
          innerleftmargin=8pt,innerrightmargin=8pt,
          innertopmargin=6pt,innerbottommargin=6pt,
          skipabove=8pt,skipbelow=8pt]{policybox}

\newmdenv[linewidth=1.2pt,linecolor=teal!55,backgroundcolor=teal!3,
          innerleftmargin=10pt,innerrightmargin=10pt,
          innertopmargin=8pt,innerbottommargin=8pt,
          skipabove=8pt,skipbelow=8pt]{tealframe}

\title{%
  \textbf{A Process Harness for Uplifting Legacy
  Workflows to Agentic BPM:
  Design and Realization in CUGA FLO}
}
\author[1,2]{Fabiana Fournier}
\author[1,2]{Lior Limonad}
\affil[1]{IBM SIL, Israel\protect\\
\texttt{\{fabiana,liorli\}@il.ibm.com}}
\affil[2]{Department of Information Systems,\protect\\
  Faculty of Computer \& Information Sciences,
  University of Haifa}
\date{June 2026}

\begin{document}
\maketitle
\thispagestyle{plain}

\begin{abstract}
We introduce the \emph{process harness}, a new mechanism for uplifting legacy
workflows into \emph{Agentic Business Process Management} (Agentic BPM) without
replacing the underlying workflow engine. A process harness places a policy-governed
agentic layer around a deterministic workflow engine, intercepting designated control
points to contribute reasoning, adaptation, and oversight while the engine retains
structural authority over the process. To define the process harness rigorously, we
develop the \textbf{Task--Decision--Flow (TDF)} model, specifying both its data
schema and its execution semantics. TDF decomposes LLM reasoning across three
policy-governed agent types: a \texttt{TaskAgent} for knowledge-intensive task
execution, a \texttt{DecisionAgent} for per-case gateway routing, and a
\texttt{FlowAgent} that governs runtime flow adaptation through a principled hook
mechanism. Each agent reasons within an explicit policy drawn from the process
\textbf{FRAME}, the aggregate policy set governing all LLM calls in the system. We
then present \textbf{CUGA FLO} as the design and implementation realization of the
TDF model, and demonstrate it on a loan approval workflow that exercises all three
agent types and hook-driven regulatory override. The process harness uniquely reconciles imperative requirements, realized through
deterministic workflow execution that enforces structural compliance, with normative
requirements, realized through policy-framed agentic autonomy invoked at designated
control points wherever the process demands it.
\end{abstract}


\section{Introduction}
\label{sec:intro}

The rapid advancement of large language models (LLMs) and the agentic frameworks
built atop them~\cite{yao2022react,wang2024survey} is prompting a fundamental
re-examination of how organizations automate and govern their business processes.
Conventional workflow systems, refined over decades of Business Process Management
(BPM) research~\cite{vdaalst2016processm}, provide rigorous structural guarantees
but offer little capacity for open-ended reasoning or adaptation to unanticipated
situations. Addressing such situations typically requires human intervention, which may trigger what practitioners call 
\emph{workarounds}: intentional deviations from the formally prescribed business process, performed by process participants to achieve process goals under perceived constraints or goal–process misalignments~\cite{outmazgin2026workarounds}.
The opportunity afforded by LLM-based agents, capable of
natural-language understanding, contextual reasoning, and tool-augmented problem
solving, is to modernize these legacy systems without displacing them: endowing
their deterministic execution backbone with a principled layer of process-aware, policy-governed
autonomy, and thereby evolving conventional automation into a new paradigm of
governed agentic intelligence, recently captured under the term of \emph{framed autonomy}~\cite{dumas2023aiaugmented,calvanese2026agenticbpm}.

Business Process Management (BPM) has for decades provided the intellectual
framework for specifying, executing, monitoring, and improving organizational
processes~\cite{vdaalst2016processm}. The standard of BPMN~2.0~\cite{bpmn2011}
gives processes a precise visual and executable semantics: tasks do work, gateways
route control flow, and sequence flows connect them. Process engines such as
Camunda\footnote{\url{https://camunda.com/}}, Flowable\footnote{\url{https://www.flowable.com/}}, IBM Business Automation Workflow\footnote{\url{https://www.ibm.com/products/business-automation-workflow}}, and SAP Signavio Process Governance\footnote{\url{https://www.signavio.com/products/process-governance/}}, compile BPMN diagrams and execute instances of them,
enforcing structural compliance by construction.

Two fundamental limitations constrain classical BPM in dynamic environments.

\paragraph{Closed-world regime.}
Every possible deviation from the nominal path must be anticipated at design time
and modeled as an explicit exception flow in the diagram. Encountering an
unanticipated situation (a regulatory change, a novel applicant profile, an
external system anomaly) leaves the process engine with no recourse other than
error and escalation. The process model is a closed world: what is not encoded
cannot happen.

\paragraph{Rigid task execution.}
Service tasks are executed by fixed software scripts that cannot adapt to
open-ended or unstructured inputs. User tasks involve human workers who bring
contextual judgment and adaptability, but introduce throughput constraints and
operational variability at scale. In neither case does the process engine itself
reason about the task context beyond what the process designer
anticipated~\cite{dumas2018fundamentals}.

Large language models (LLMs) offer a compelling complement: open-ended reasoning,
natural-language understanding, and the ability to synthesize plans from ambiguous
inputs. The emergent paradigm of LLM agents~\cite{yao2022react,wang2024survey}
extends this to tool use and multi-step problem solving. Deploying an LLM \emph{within} individual tasks (the AI-augmented paradigm)
preserves the process topology but leaves the overall control flow rigid and
incapable of runtime adaptation. Deploying an LLM \emph{as a planner} that
reads a BPMN description and generates an execution plan does so outside the
process engine, with no mechanism to guarantee that the execution conforms to
the model, respects audit requirements, or enforces compliance policies.

We argue that the resolution to this tension is not a choice between structured
rigidity and open-ended flexibility, but a new class of system that combines both:
an agent that \emph{oversees} the process graph and is able to recommend in-flight
process interventions while being governed by respective policies. Such a system was recently envisioned in the Agentic BPM
manifesto~\cite{calvanese2026agenticbpm}, which positions agents as first-class
functional entities within business processes. However, the manifesto remains
agnostic to any concrete implementation. In this work, we introduce CUGA FLO as an agentic \emph{process harness}: a new
concept in the BPM arena, coined in analogy to the recently introduced notion of
\emph{agent harness}~\cite{meng2026agentharness}, denoting a mechanism that engages
LLM agents with the execution of a workflow engine as an overlay layer that enables an
existing process to behave agentically without rewriting the process engine. We developed its conceptual underpinnings as a first realization in adherence to the manifesto principles and its mappings to concrete software entities. CUGA FLO enables any
existing workflow system to be transformed into a fully agentic one, with a clean
separation between policy-governed agentic oversight and workloads on one side, and deterministic
workflow execution on the other, the latter remaining the responsibility of the
workflow engine CUGA FLO integrates with.

\paragraph{Agentic BPM: normative and imperative requirements.}
CUGA FLO conforms to the Agentic BPM view, which calls for the combination of
both \emph{normative} and \emph{imperative} requirements about a business
process~\cite{calvanese2026agenticbpm,dumas2023aiaugmented}. The normative layer is
formalised in CUGA FLO as the \textbf{FRAME} $\mathcal{F}$, the aggregate set of
policy documents that bounds every LLM call in the system
(Section~\ref{sec:tdf}).
The \textbf{imperative} part is captured in the process model: an explicit,
executable graph of tasks, gateways, and sequence flows that the workflow engine
compiles and enforces at runtime.
The \textbf{normative} part is expressed through the set of \textbf{policies} disseminated
in the system among the three agent types: \texttt{TaskAgent} policies govern
what a task may produce, \texttt{DecisionAgent} policies govern which path a
gateway may select, and hook policies govern what structural interventions are
permissible at a given flow point.
Together, these two layers give CUGA FLO both the structural guarantees of a
process engine and the adaptive reasoning of a policy-bounded LLM system.
This is realised through three agent types, \texttt{TaskAgent},
\texttt{DecisionAgent}, and \texttt{FlowAgent}, each governed by its own FRAME
policy. In addition, per-process \texttt{action\_permissions} constitute a second
governance layer, formalized in Section~\ref{sec:tdf} as the access control
function $\phi$: it provides an explicit enumeration of which structural
intervention types the FlowAgent may and may not apply, declared in the process
configuration independently of the hook policies, and is also enforced in the execution semantics.

ConfigUrable Generalist Agent (CUGA)\footnote{\url{https://github.com/cuga-project/cuga-agent}}, is an open-source generalist agent harness for the enterprise, supporting complex task execution on web and APIs, OpenAPI/MCP integrations, composable architecture, reasoning modes, and policy-aware features. Its core primitive is \texttt{CugaAgent}, a central execution hub that selectively triggers planning, tool shortlisting, and context summarisation only when the task demands it, while maintaining direct access to policies and memory to ensure every interaction remains personalised and policy-compliant. \texttt{CugaSupervisor} serves as the orchestration layer for high-complexity tasks: it delegates tasks to specialized sub-agents, mixes local agents with remote A2A agents, and passes data between them. CUGA FLO builds on these primitives: each \texttt{TaskAgent} and \texttt{DecisionAgent} wraps a \texttt{CugaAgent}, and the \texttt{FlowAgent} itself also wraps a \texttt{CugaAgent} for its internal hook reasoning, while presenting a unified \texttt{invoke()} interface to a parent \texttt{CugaSupervisor}.

\paragraph{Why Agentic BPM?}
Real-world business processes exhibit what practitioners call a \emph{spaghetti}
distribution of execution traces~\cite{vdaalst2016processm}: a small number of
frequent, well-structured variants account for the bulk of automated case handling,
while a long tail of less frequent, specialised, or context-specific variants
is routinely handled outside the automated process, off the nominal control flow. Conventional BPM automation handles the frequent
variants reliably but leaves the long tail to manual intervention. CUGA FLO extends
automation coverage to this long tail through governed agentic intervention: because
hook policies reason about cases rather than enumerate paths, rare variants are handled
by the same mechanism as frequent ones, without requiring programmatic modifications to the process model.

At an organisational level, CUGA FLO acts as the missing modernisation apparatus
that enables any existing workflow system to be transformed incrementally into a
fully agentic one: from agent-augmented BPM, in which agents assist with specific
activities, to full Agentic BPM, in which agents govern the process end-to-end.
The transformation is gradual and reversible: any of the three process harness autonomy
levels (Section~\ref{sec:paradigm}) can be activated or deactivated per process
without changing the underlying engine.

\paragraph{Contributions.}
This paper makes the following contributions:

\begin{enumerate}[label=\textbf{C\arabic*}., noitemsep]
  \item We introduce the \textbf{Process Harness} paradigm as the means to uplift
        any conventional workflow system into a full Agentic BPM
        system~\cite{calvanese2026agenticbpm}, characterised by three key
        properties (structural conformance, policy accountability, and runtime
        adaptability) that distinguish it from classical BPM and LLM-as-planner
        approaches (\S\ref{sec:paradigm}).
  \item We introduce the \textbf{Task--Decision--Flow (TDF) model} as the conceptual underpinning for the process harness, a principled
        decomposition of LLM reasoning across three agent types: \texttt{TaskAgent},
        \texttt{DecisionAgent}, and \texttt{FlowAgent} (\S\ref{sec:tdf}).
  \item We instantiate the \textbf{FRAME} concept~\cite{dumas2023aiaugmented} as
        the aggregate policy set $\mathcal{F}$ governing a TDF process, and show how
        partitioning it across three agent types enforces separation of concerns at
        the LLM level (\S\ref{sec:tdf}).
  \item We present \textbf{CUGA FLO} and its hook mechanism as the first
        realization of the process harness, enabling runtime flow adaptation and a Model Context Protocol (MCP) bridge that decouples the
        reasoning performed by all process harness agents from the deterministic execution of a replaceable workflow engine
        (\S\ref{sec:system}).
  \item We demonstrate the model on a small illustrative example of a \textbf{loan approval workflow} that
        exercises all three agent types and a regulatory override hook
        (\S\ref{sec:casestudy}).
\end{enumerate}

\section{A Process Harness Paradigm}
\label{sec:paradigm}

A \emph{process harness} is an agentic layer that wraps an existing workflow
system without replacing it. The underlying engine retains ownership of the
process model and drives execution. The process harness intercepts designated control
points (tasks, gateways, and sequence flows) and engages policy-governed LLM
agents at each one, contributing reasoning, adaptation, and oversight. This
paradigm is the mechanism for uplifting any conventional workflow system into a
full Agentic BPM system~\cite{calvanese2026agenticbpm}: the transformation is
additive and reversible, leaving the existing execution infrastructure intact while
placing a principled agentic reasoning layer around it.

Thus we define a process harness as follows:
\begin{definition}
    A \texttt{process harness} is a Task--Decision--Flow-based agentic layer placed around a deterministic workflow engine, enabling legacy workflows to be uplifted into Agentic Business Process Management through framed reasoning, interventions, and runtime adaptations without altering the underlying workflow semantics.
\end{definition}

Two fundamental principles, drawn from the Agentic BPM
manifesto~\cite{calvanese2026agenticbpm}, govern every agent in the process harness.
\textbf{Process awareness}: each agent receives, at the moment of engagement, the
process model, the current execution state, and the history of prior steps, so its
reasoning is fully grounded in the live context of the running process.
\textbf{Framing}: each agent reasons within an explicit, human-readable policy that
defines what it may produce, ensuring agentic behavior is neither arbitrary nor
opaque.

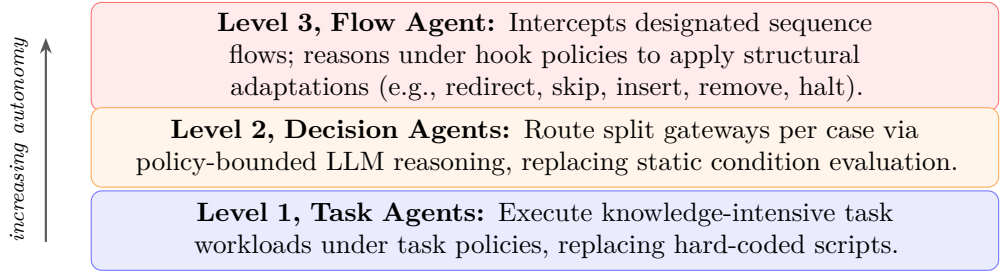
\begin{figure}[H]
\centering
\begin{tikzpicture}[font=\small]

\node[draw=red!55, fill=red!8, rounded corners=4pt,
      minimum width=12cm, minimum height=1.05cm, text width=11.5cm,
      text centered] (l3) at (0, 2.3)
      {\textbf{Level~3, Flow Agent:} Intercepts designated sequence flows; reasons
       under hook policies to apply structural adaptations (e.g., redirect, skip, insert,
       remove, halt).};

\node[draw=orange!65, fill=orange!8, rounded corners=4pt,
      minimum width=12cm, minimum height=1.05cm, text width=11.5cm,
      text centered] (l2) at (0, 1.1)
      {\textbf{Level~2, Decision Agents:} Route split gateways per case via
       policy-bounded LLM reasoning, replacing static condition evaluation.};

\node[draw=blue!55, fill=blue!8, rounded corners=4pt,
      minimum width=12cm, minimum height=1.05cm, text width=11.5cm,
      text centered] (l1) at (0, 0)
      {\textbf{Level~1, Task Agents:} Execute knowledge-intensive task workloads
       under task policies, replacing hard-coded scripts.};

\draw[-{Stealth[length=6pt]}, thick, black!65] (-6.6, -0.15) -- (-6.6, 2.55);
\node[rotate=90, font=\scriptsize\itshape, anchor=south] at (-6.7, 1.2)
      {increasing autonomy};

\end{tikzpicture}
\caption{Three composable levels of agentic autonomy enabled by a process harness.
Each level is independently activatable per process. All three share the same
process knowledge upon engagement.}
\label{fig:autonomy-levels}
\end{figure}

A \emph{process harness} is a mechanism that resides between a workflow execution engine and an agentic reasoning layer, mediating their interaction. It is characterized as follows:

\begin{enumerate}[noitemsep, label=(\roman*)]
  \item The process harness does not own or execute the process graph. A
        \texttt{WorkflowEngine} retains that responsibility. The process harness engages
        LLM-backed agents at designated control points, passing each agent the
        current process state, execution history, and model as input context.
  \item Every agent reasoning call is bounded by an explicit, human-readable
        policy.
  \item The process harness enables three composable levels of agentic autonomy as illustrated in~\Cref{fig:autonomy-levels}, each
        independently activatable per process:
        \begin{itemize}[noitemsep]
          \item \emph{Level~1: Task delegation}: task agents handle
                knowledge-intensive task workloads, replacing hard-coded scripts
                with policy-governed LLM execution.
          \item \emph{Level~2: Routing delegation}: decision agents handle
                gateway routing decisions, replacing static rules with per-case
                LLM reasoning bounded by a decision policy.
          \item \emph{Level~3: Flow adaptation}: a flow agent intercepts
                designated sequence flows and reasons under hook policies to recommend
                structural interventions to the running process.
        \end{itemize}
  \item Process knowledge, comprising the process model, current execution state, and
        execution history, is exchanged between the \texttt{WorkflowEngine} and
        the process harness at every selected control point, sustaining full process awareness in
        every agent reasoning call.
\end{enumerate}

Control points are the workflow engine's designated callback invocations: the engine
pauses process execution upon reaching a control point, calls into the process harness
to engage the respective agent for task fulfillment, routing decisions, or hook
interventions, awaits a recommendation response, applies the recommendation, and then resumes execution accordingly.

\begin{figure}[H]
\centering
\begin{tikzpicture}[node distance=1.5cm and 2.2cm, font=\small]

\node (xl) at (-0.7, 2.9) [rotate=90, anchor=center, font=\footnotesize\itshape]
      {Structural Conformance};
\node (yl) at (4, -0.7) [anchor=center, font=\footnotesize\itshape]
      {Runtime Adaptability};

\node[draw=black!25, fill=red!5, minimum width=3.8cm, minimum height=2.8cm,
      rounded corners=4pt] at (1.4, 1.2) {};
\node[draw=black!25, fill=green!6, minimum width=3.8cm, minimum height=2.8cm,
      rounded corners=4pt] at (6.2, 1.2) {};
\node[draw=black!25, fill=yellow!5, minimum width=3.8cm, minimum height=2.8cm,
      rounded corners=4pt] at (1.4, 4.2) {};
\node[draw=black!25, fill=blue!5, minimum width=3.8cm, minimum height=2.8cm,
      rounded corners=4pt] at (6.2, 4.2) {};

\draw[-{Stealth}, thick, black!50] (-0.3,0) -- (8.2,0) node[right,font=\footnotesize]{High};
\draw[-{Stealth}, thick, black!50] (-0.3,0) -- (-0.3,5.8) node[above,font=\footnotesize]{High};

\node[align=center, font=\footnotesize] at (1.4, 4.2)
      {\textbf{Classical BPM}\\High conformance,\\low adaptability};
\node[align=center, font=\footnotesize\color{red!60!black}] at (6.2, 1.2)
      {\textbf{LLM-as-Planner}\\Low conformance,\\high adaptability};
\node[align=center, font=\footnotesize] at (1.4, 0.6)
      {\textbf{Rule-based pipeline}\\Rigid, no LLM};
\node[align=center, font=\footnotesize\bfseries\color{blue!60!black}] at (6.2, 4.2)
      {$\bigstar$ \textbf{CUGA FLO}\\High conformance,\\high adaptability};

\draw[black!40] (4, 0.05) -- (4, -0.05) node[below,font=\scriptsize]{};
\draw[black!40] (0.05, 3) -- (-0.05, 3) node[left,font=\scriptsize]{};

\end{tikzpicture}
\caption{Positioning of a process harness (realized as CUGA FLO) relative to
existing approaches along structural conformance and runtime adaptability axes.}
\label{fig:positioning}
\end{figure}
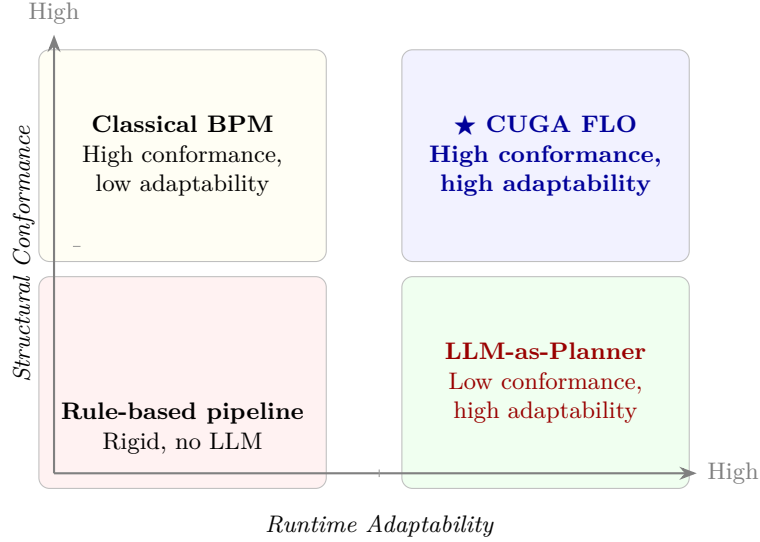

The process harness paradigm resolves the tension between structural rigidity and
open-ended flexibility by separating two concerns. \emph{Process topology} is
fixed: the set of permissible activities and their ordering constraints is
specified in the process model and cannot be violated. \emph{Process adaptation} is
dynamic: permissible deviations are governed by the aggregate policy set (the FRAME,
Section~\ref{sec:tdf}) and enacted at runtime by the flow agent reasoning under
that FRAME. Figure~\Cref{fig:positioning} positions the proposed process harness, realized as CUGA FLO, relative to existing approaches along the dimensions of structural conformance and runtime adaptability. A broader comparison with related paradigms is presented in Section~\ref{sec:comparison}.

This is not exception handling in the classical sense. Classical exception handling
encodes specific deviation paths devised during design time. A process harness acts as an \emph{open-world adaptation layer}:
the set of situations that can be handled is the set of situations the FRAME
policies can reason about, which is unbounded by design. Moreover, at every
gateway, a decision agent replaces static condition evaluation with per-case LLM
reasoning, extending adaptability beyond hook-driven structural interventions to
the routing logic itself. This makes the system highly resilient and inherently adaptable in response to unforeseen events and conditions during overall process execution~\cite{marron2020unexpected}.

The formal model underlying a process harness is the Task--Decision--Flow (TDF)
model, presented in Section~\ref{sec:tdf}. CUGA FLO is the first realization of
this model; its concrete design and implementation are described in
Section~\ref{sec:system}.

\section{The Task--Decision--Flow Model}
\label{sec:tdf}

The Task-Decision-Flow (TDF) model assigns LLM-agent reasoning to three distinct roles, each bound to a
specific class of process-model element or execution concern, each with a distinct
role and FRAME policy. The TDF process harness wraps
an underlying process model and lifts it into an agentic execution layer.

\subsection{The TDF Schema}
\label{sec:tdf-schema}

The TDF model is defined with respect to a given process or workflow model. At the
most fundamental level, the process model is represented as a directed graph:
\[
  Pm = (V, E)
\]
where $V$ is the set of process nodes and $E \subseteq V \times V$ is the set of
directed process flows between nodes. Each node and each edge is associated with a
unique element identifier. The process model is not required to be acyclic, where cycles
may represent iterative or looping behaviour.

The process harness abstracts heterogeneous workflow notations (e.g., BPMN and EPC) into a Task--Decision--Flow graph comprising task nodes, gateway nodes, and directed flow edges, thereby enabling a unified process model.

The nodes in $V$ may represent process tasks, gateways, start nodes, or end nodes:
\[
  V = V_{start} \cup V_{end} \cup T \cup G_{split} \cup G_{join}
\]
where $V_{start}$ is the set of start nodes, $V_{end}$ is the set of end nodes,
$T$ is the set of task nodes, $G_{split}$ is the set of split gateway nodes,
comprising conditional split gateways $G_{split}^{cond}$ (whose outbound routing
depends on a condition evaluated against process state) and unconditional gateways
(parallel splits that activate all outbound flows simultaneously), and
$G_{join}$ is the set of join gateway nodes. Start nodes have only outbound flows,
end nodes have only inbound flows, split gateways determine one or more outbound
flows, and join gateways merge or synchronize inbound flows.

Each task $t \in T$ is specified by a set of goals and a set of attributes:
\[
  Goals(t) \subseteq GoalStmt
  \qquad
  Attrs(t) \subseteq Attr
\]
where $GoalStmt$ is the set of possible goal statements and $Attr$ is the set of
possible task attributes. The output of executing a task may be viewed as assigning
values to some or all of its attributes.

Each conditional split gateway $g \in G_{split}^{cond}$ is specified by a condition statement:

\[
Cond(g) = \{p_1, \ldots, p_n\}
\]

where each $p_i$ is a logical proposition associated with an outbound flow of $g$. The evaluation of these propositions determines which outbound flow(s) are selected during execution.

In addition to the process model, the process has a state and a history. The process
state records the execution status of each task:
\[
  Ps : T \rightarrow \{\mathsf{pending}, \mathsf{active}, \mathsf{hold},
  \mathsf{completed}, \mathsf{halted}\}
\]
where \textsf{pending} means the task awaits execution, \textsf{active} means it is
currently executing, \textsf{hold} means it was started and is waiting to resume,
\textsf{completed} means it has finished, and \textsf{halted} means it was terminated before completion. The process history $Ph$ is the
ordered trace of past executed elements up to the current process state.

The collective knowledge about the process is:
\[
  P = (Pm, Ps, Ph)
\]
This knowledge captures the structure of the process, the current state of its
execution, and the process history.

With respect to a given process knowledge $P$, a TDF model defines a process
harness as a tuple:
\[
\mathbf{TDF}_P = (TAs, DAs, FAs)
\]
where $TAs$ is a set of task agents, $DAs$ is a set of decision agents, and $FAs$
is a set of flow agents. These three kinds of agents form an agentic process harness around
the process model: task agents map to tasks, decision agents map to split gateways,
and flow agents map to selected process flows (namely, hooks).

A \textbf{task agent} is defined as:
\[
  a = (t, C_T, P, Ts, \lambda)
\]
where $t \in T$ is the task the agent is assigned to, $C_T$ is the set of policies
governing the task, $P$ is the process knowledge, $Ts$ is the set of tools
available to the agent, and $\lambda$ is the internal LLM-based agent that performs
the actual sense--reason--act loop. The quality criterion for a task agent is policy
compliance and correctness of domain output.

A \textbf{decision agent} is defined as:
\[
  d = (g, C_D, P, Ts \cup \{cond\_eval\}, \lambda)
\]
where $g \in G_{split}^{cond}$ is the conditional split gateway the agent is assigned to, $C_D$ is the
set of policies governing the decision, $P$ is the process knowledge, $Ts$ is the
set of tools available to the agent, $cond\_eval$ is a mandatory
condition-evaluation tool, and $\lambda$ is the internal LLM-based agent. The
special role of $cond\_eval$ is to ensure that the gateway condition is explicitly
evaluated before the agent reasons about the outbound routing decision. The quality
criterion for a decision agent is determinism and correctness of the selected
outbound flows (exactly one for exclusive gateways; one or more for inclusive
gateways).

A \textbf{flow agent} is defined as:
\[
  f = (H, Is, \phi, C_F(h_k), P, Ts, \mu, \lambda)
\]
where $H \subseteq E$ is a set of hooks (selected process flows on which the agent
is active), $Is$ is the set of intervention action primitives available to the
agent, $\phi: Is \to \{\mathsf{permitted}, \mathsf{prohibited}\}$ is a per-process
access control function that governs which intervention types the agent may instruct
(elaborated in Section~\ref{sec:tdf-semantics}), $C_F(h_k)$ is the set of policies
governing the flow interventions, partitioned by the hooks $h_k \in H$, $P$ is the process
knowledge (where $E$ denotes the sequence flows of the process model $Pm$, i.e., the edges of $P$, and $H \subseteq E$), $Ts$ is the set of tools available to the agent, $\mu : V_{id} \to
TAs \cup DAs$ is a mapping function from process element identifiers to their
corresponding task or decision agents (i.e., some task agent $a \in TAs$ or some decision agent $d \in DAs$ in $TDF_P$), and $\lambda$ is the internal LLM-based
agent. Examples of intervention primitives include continue, skip task, swap tasks,
remove task, add task, or terminate. The quality criterion for a
flow agent is FRAME conformance and auditability of interventions.

Each of the three TDF agent types wraps an internal LLM-based agent:
\[
  \lambda = (M, Ts, sense, reason, act)
\]
where $M$ is the agent memory, $Ts$ is the set of tools (which maps directly to the
tools assigned to the wrapping agent), $sense$ is the sensing function, $reason$ is
the reasoning function, and $act$ is the acting function. The set of tools $Ts$ may, for any of the three agent types, include designated tools to interact with the user (e.g., an \texttt{ask\_user} tool), enabling the agent to solicit input when needed to fulfill its reasoning. The TDF model does not
specify the internal implementation of the LLM agent; it specifies how the agent is
contextualized by the process harness and how its behaviour is constrained by the
role it plays in the process. The model is intentionally agnostic to the capability
profile of the underlying agents, as different agents may exhibit different levels of
fitness and skills for a given mission.

\subsection{TDF Execution Semantics}
\label{sec:tdf-semantics}

The operation semantics of the TDF model describes how the three agent types behave
at runtime. Each TDF agent wraps an internal LLM-based agent with a
sense--reason--act loop. The loop is always informed by two kinds of knowledge:
process awareness and framing.

\emph{Process awareness} is provided by $P = (Pm, Ps, Ph)$ and includes the process
model, the current process state, and the process history (the imperative knowledge). Through process awareness,
the agent can reason about the task it is assigned to, the surrounding graph
structure, the state of other tasks, gateway conditions, and the current execution
context.

\emph{Framing} is provided by the policy sets that define the
constraints, preferences, or norms that the agent is expected to follow (the normative knowledge, typically articulated in deontic language form). Thus, every
TDF agent is both process-aware and policy-framed. The TDF model relies on the
internal LLM-based agent to use this context, together with its tools, to reason and
act in a way that is adequate for the prompt and aligned with the supplied policies.

For a task agent $a = (t, C_T, P, Ts, \lambda)$, the sensing step assigns the agent
memory with the relevant process knowledge and policies: the agent becomes aware of
the process model, the current process state and history, and the policy frame under
which it should operate. The reasoning step then prompts the internal LLM-based
agent to fulfil the goals of the wrapped task $t$, using the tools in $Ts$ to infer
an execution plan. The acting step executes this plan and captures the result as an
assignment of values to the attributes of the task:
\[
  TaskRun(a) = \alpha_t
\]
where $\alpha_t$ denotes the task output over the attributes of $t$. Thus, the role
of a task agent is to transform the goals and context of a task into an executable
plan and then into task-level outputs.

For a decision agent $d = (g, C_D, P, Ts \cup \{cond\_eval\}, \lambda)$, the sensing
step is similar: the agent memory is assigned with process knowledge and policy
framing. However, the decision agent has a mandatory operational constraint. Before
reasoning about which outbound flows should be selected, it must invoke the
condition-evaluation tool:
\[
  cond\_eval(Cond(g), P)
\]
The result of this evaluation is added to the agent memory. Only after this step may
the internal LLM-based agent reason about the routing decision, using the gateway
condition result, the process context, the policy frame, and the available tools to
infer one or more outbound flows from $g$. The acting step emits the selected routing
decision:
\[
  DecisionRun(d) = R_g
\]
where $R_g$ is a non-empty set of outbound flows from the conditional split gateway. Thus, the
role of a decision agent is to provide controlled, condition-informed routing at
split gateways.

For a flow agent $f = (H, Is, \phi, C_F(h_k), P, Ts, \mu, \lambda)$, the sensing
step assigns the agent memory with process knowledge and policy framing. The flow
agent bears two distinct responsibilities. When execution reaches a hooked flow
$h_k \in H$, a hook callback issued by the workflow engine triggers the flow agent to initiate hook reasoning: the internal agent $\lambda$ is
invoked under the hook policy $C_F(h_k)$ together with the current process knowledge
to evaluate the appropriate intervention. Upon reaching an agentic task or a conditional split gateway, a corresponding callback by the workflow engine triggers the flow agent to engage the respective task agent or decision agent. In response, the flow agent delegates task
execution and routing responsibilities to the appropriate agents via the mapping
$\mu$: a task or decision agent is either identified explicitly through $\mu$ (based
on the process element identifier), or resolved dynamically by $\lambda$ from a
supplied goal or condition statement when no explicit mapping exists. To this end,
an implementation may incorporate agent-selection, skill-routing, or
capability-composition mechanisms that dynamically allocate goal missions to the most
suitable agents and augment them with task-specific skills, tools, or knowledge
resources as needed. Such mechanisms are orthogonal to the TDF abstraction and
concern the optimization of mission execution rather than the definition of
process-level agentic roles.

During hook reasoning, the internal LLM-based agent considers the hooked flows, the
process state and history, the policies, and the available tools in order to infer
an intervention from the available primitives $Is$. Once the reasoning step produces
a candidate intervention $i_f \in Is$, the process harness evaluates $\phi(i_f)$ as
a second governance layer operating externally to the LLM loop. Only if
$\phi(i_f) = \mathsf{permitted}$ does the acting step proceed to emit the
instruction. If $\phi(i_f) = \mathsf{prohibited}$, the instruction is suppressed
and the process harness substitutes a safe default action (e.g., \textsc{continue}). The acting step then emits the
instruction for the permitted intervention:
\[
  FlowRun(f) = i_f
\]
where $i_f \in Is$. Thus, the role of a flow agent is to support process-level
oversight by recommending interventions such as continuing execution, skipping a task,
swapping tasks, pausing the process, or terminating it when deemed necessary.

\paragraph{Access permission governance.}
Beyond the FRAME, the TDF model provides a second governance layer through the
per-process access control function $\phi$ introduced in the flow agent schema. At
the schema level, $\phi: Is \to \{\mathsf{permitted}, \mathsf{prohibited}\}$
partitions the available intervention primitives into those the FlowAgent may recommend
and those it may not, declared independently of any hook policy. At the semantic
level, $\phi$ is evaluated by the process harness after every hook reasoning cycle
and before any instruction is emitted to the workflow engine. This two-layer
architecture separates concerns clearly: the FRAME bounds what the LLM may reason
about and conclude, while $\phi$ bounds what the process harness may actually trigger the workflow engine to act upon, irrespective
of what reasoning produces. Moreover, both the $\phi$ permissions and the policies may be introduced or revised at any point
in time, with changes taking effect immediately on all subsequent agent reasoning
calls within running process instances.

The TDF model adheres to the two fundamental principles of Agentic
BPM~\cite{calvanese2026agenticbpm}. \textbf{Process awareness} is ensured by making
the process knowledge $P = (Pm, Ps, Ph)$ visible to every agent upon engagement:
each receives the current state, history, and model as part of its input context,
enabling reasoning that is informed by the full process trajectory.
\textbf{Framing} is ensured by associating each agent with an explicit policy
drawn from the process FRAME $\mathcal{F}$~\cite{calvanese2026agenticbpm,dumas2026abpms}:
\[
  \mathcal{F} = \bigl\{C_T(t_i)\bigr\}_{i=1}^n \;\cup\;
                \bigl\{C_D(d_j)\bigr\}_{j=1}^m \;\cup\;
                \bigl\{C_F(h_k)\bigr\}_{k=1}^p
\]
where $C_T$, $C_D$, $C_F$ are task, decision, and hook policies respectively.
No LLM call occurs outside a policy boundary, where the policy determines what the agent may reason about and produce.

\section{CUGA FLO System Architecture}
\label{sec:system}

\subsection{Overview}

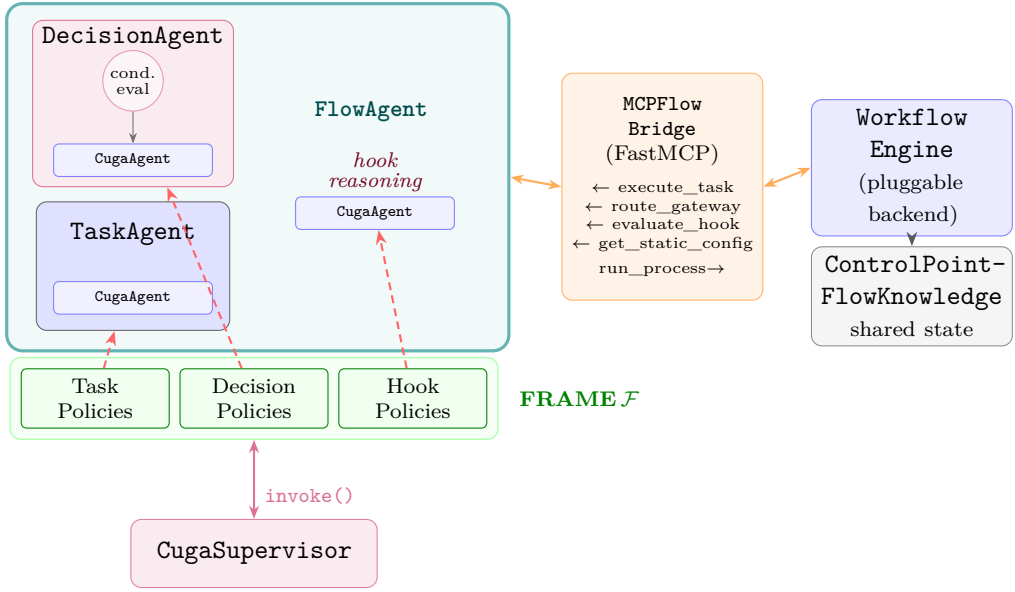
\begin{figure}[H]
\centering
\begin{tikzpicture}[node distance=1.1cm and 1.6cm, font=\small]

\coordinate (fa@bl) at (-1.0,1.65);
\coordinate (fa@tr) at (5.5,6.10);
\begin{scope}[on background layer]
  \node[frame, fit=(fa@bl)(fa@tr), inner sep=2pt] (fabox) {};
\end{scope}
\node[font=\footnotesize\bfseries\color{teal!60!black}] at (3.75,4.75)
      {\texttt{FlowAgent}};

\node[agent, minimum height=2.2cm] (da) at (0.6, 4.85) {};
\node[font=\small, align=center] at (0.6, 5.72) {\texttt{DecisionAgent}};
\node[circle, draw=purple!40, fill=purple!4, minimum size=0.8cm,
      font=\tiny, align=center, inner sep=1pt] (ceval) at (0.6, 5.15)
      {cond.\\[-1pt]eval};
\node[draw=blue!50, fill=blue!6, rounded corners=2pt,
      minimum width=2.1cm, minimum height=0.4cm, font=\tiny, align=center]
      (da@ca) at (0.6, 4.10) {\texttt{CugaAgent}};
\draw[-{Stealth[length=4pt]}, thin, black!55] (ceval) -- (da@ca);

\node[task, minimum height=1.7cm] (ta) at (0.6, 2.7) {};
\node[font=\small, align=center] at (0.6, 3.12) {\texttt{TaskAgent}};
\node[draw=blue!50, fill=blue!6, rounded corners=2pt,
      minimum width=2.1cm, minimum height=0.4cm, font=\tiny, align=center]
      at (0.6, 2.28) {\texttt{CugaAgent}};

\node[font=\scriptsize\itshape, color=purple!60!black, align=center]
      at (3.8, 3.95) {hook\\[-2pt]reasoning};
\node[draw=blue!50, fill=blue!6, rounded corners=2pt,
      minimum width=2.1cm, minimum height=0.4cm, font=\tiny, align=center]
      (hm) at (3.8, 3.4) {\texttt{CugaAgent}};

\node[policy] (pt) at (0.1, 0.95) {Task\\Policies};
\node[policy] (pd) at (2.2, 0.95) {Decision\\Policies};
\node[policy] (ph) at (4.3, 0.95) {Hook\\Policies};
\begin{scope}[on background layer]
  \node[draw=green!40,fill=green!4,rounded corners=4pt,
        fit=(pt)(pd)(ph),inner sep=4pt] (framebox) {};
\end{scope}
\node[font=\scriptsize\bfseries\color{green!50!black}, anchor=west]
      at ([xshift=4pt]framebox.east) {FRAME\,$\mathcal{F}$};
\draw[hookflow] (pd) -- (da);
\draw[hookflow] (ph) -- (hm);
\draw[hookflow] (pt) -- (ta);

\node[draw=orange!60, fill=orange!10, rounded corners=4pt,
      minimum width=2.5cm, minimum height=3.0cm, text width=2.4cm,
      text centered, font=\scriptsize] (bridge) at (7.6,3.75)
      {\texttt{MCPFlow}\\\texttt{Bridge}\\\scriptsize(FastMCP)\\[5pt]
       \tiny $\leftarrow$ execute\_task\\\tiny $\leftarrow$ route\_gateway\\\tiny $\leftarrow$ evaluate\_hook\\\tiny $\leftarrow$ get\_static\_config\\\tiny run\_process$\rightarrow$ };

\node[draw=blue!50, fill=blue!8, rounded corners=4pt,
      minimum width=2.6cm, minimum height=1.0cm, text centered, text width=2.4cm]
      (engine) at (10.9,4.0) {\texttt{Workflow}\\\texttt{Engine}\\\scriptsize(pluggable backend)};
\node[draw=black!45, fill=gray!8, rounded corners=4pt,
      minimum width=2.6cm, minimum height=0.9cm, text centered, text width=2.4cm]
      (fs) at (10.9,2.3) {\texttt{ControlPoint-\\FlowKnowledge}\\\scriptsize shared state};
\draw[flow] (engine) -- (fs);

\draw[{Stealth[length=5pt]}-{Stealth[length=5pt]}, thick, orange!65]
      (fabox.east) -- (bridge.west);
\draw[{Stealth[length=5pt]}-{Stealth[length=5pt]}, thick, orange!65]
      (bridge.east) -- (engine.west);

\node[draw=purple!50, fill=purple!8, rounded corners=4pt,
      minimum width=3.2cm, minimum height=0.9cm, text centered, text width=3cm]
      (sup) at (2.2, -1.1) {\texttt{CugaSupervisor}};

\draw[{Stealth[length=5pt]}-{Stealth[length=5pt]}, thick, purple!55]
      (framebox.south) -- (sup) node[pos=0.7, right, font=\scriptsize]{\texttt{invoke()}};

\end{tikzpicture}
\caption{CUGA FLO architecture. The reasoning side (left), comprising the \texttt{FlowAgent}
with its \texttt{DecisionAgent} and \texttt{TaskAgent} sub-agents (each backed by a
\texttt{CugaAgent}) and the FlowAgent's own \texttt{CugaAgent} for hook reasoning, all governed by the FRAME $\mathcal{F}$, is decoupled from the execution
side (right) by the \texttt{MCPFlowBridge}. The bridge exposes the FlowAgent's
callback listeners (\texttt{execute\_task}, \texttt{route\_gateway},
\texttt{evaluate\_hook}) and the engine's
\texttt{run\_process} tool. A replaceable \texttt{WorkflowEngine} drives execution over
\texttt{ControlPointFlowKnowledge} and calls back into the FlowAgent at each control point. The FlowAgent presents a unified
\texttt{invoke()} interface to a \texttt{CugaSupervisor} agent.}
\label{fig:architecture}
\end{figure}

CUGA FLO implements the TDF model as a Python library that separates
\emph{policy-aware reasoning} from \emph{process execution} (see~\Cref{fig:architecture}). The reasoning side
(the \texttt{FlowAgent} and its \texttt{TaskAgents} and \texttt{DecisionAgents}) is decoupled from the
execution side (a \texttt{WorkflowEngine} that drives the BPMN process graph) by
the \texttt{MCPFlowBridge}, a FastMCP server that mediates all communication between
them. The FlowAgent registers the following callback listeners as MCP tools on the bridge:
\texttt{execute\_task}, \texttt{route\_gateway}, and \texttt{evaluate\_hook}. Any workflow engine integrating with CUGA FLO calls
those tools at every control point.

\subsection{The FlowAgent: Process-Aware Meta-Agent}

The \texttt{FlowAgent} is the central reasoning component of CUGA FLO. All three TDF
agent types are process-aware: each receives, upon engagement by the
\texttt{WorkflowEngine}, the current process state, execution history, and relevant
process model summary as part of its \texttt{ControlPointFlowKnowledge} (i.e., the process knowledge). What distinguishes
the \texttt{FlowAgent} is its \textbf{structural intervention authority}: it is the
sole component in the system permitted to suggest altering the nominal process path,
always under hook policy and within declared \texttt{action\_permissions}. Such
modifications are suggested by the FlowAgent and carried out by the workflow engine.
It also has
\textbf{global process visibility}: during hook reasoning it receives the full set
of remaining unexecuted task nodes, enabling the hook LLM to reason about
consequences for the remainder of the case.

Crucially, the FlowAgent does not execute the process graph. At startup, it
registers its callback listeners (\texttt{execute\_task}, \texttt{route\_gateway},
\texttt{evaluate\_hook}, \texttt{get\_static\_config}) on the \texttt{MCPFlowBridge}.
The workflow engine, during its build phase, fetches process configuration via
\texttt{get\_static\_config} (task and gateway identifiers, hook definitions, flow
conditions, and declared action permissions), using it to integrate the control points
at which it will trigger the callbacks into CUGA FLO. At runtime a \texttt{WorkflowEngine} drives execution and invokes these tools
at each control point, passing a \texttt{ControlPointFlowKnowledge} that embeds the
full process state, execution history, model summary, and task instruction. Each
reasoning call is thereby self-contained and stateless from the engine's perspective.

\paragraph{Hook-based runtime adaptation.}
Hooks are the central mechanism for runtime adaptation in CUGA FLO. In the general
model, \textbf{a hook is an annotation over a process sequence flow}: exactly one
hook may be attached to any given flow. When execution reaches an annotated
transition, the workflow engine intercepts it and issues a callback to CUGA FLO to handle it, and the FlowAgent reasons, using the current
process knowledge and the hook's policy $C_F(h_k)$, about how execution should
proceed before the target node is entered. Each hook carries a \texttt{location}
(the flow identifier it intercepts) and a \texttt{policy} $C_F(h_k)$ drawn from the
process FRAME. The FlowAgent provides to the reasoning call the hook policy, the
current process state (execution path, variables, task results), and, uniquely, the
dictionary of remaining unexecuted tasks with their identifiers and names.

The hook LLM reasons under $C_F(h_k)$ and returns a structured action with one of
seven intervention types listed in~\Cref{tab:actions}.

\begin{table}[H]
\centering
\small
\begin{tabularx}{\linewidth}{@{}llX@{}}
\toprule
\textbf{Action} & \textbf{Structural Effect} & \textbf{Semantics} \\
\midrule
\textsc{continue}
  & None (proceed normally)
  & Policy is satisfied; follow the nominal edge \\[2pt]
\textsc{skip\_node}
  & Skip the immediate next node
  & The next node records itself as skipped and control passes to its
    successor \\[2pt]
\textsc{skip\_to}
  & Jump to any named task node
  & Bypasses all intermediate nodes; target is any remaining task \\[2pt]
\textsc{swap\_nodes}
  & Exchange two nodes
  & Redirect to \texttt{node\_b} when \texttt{node\_a} was next, or to
    \texttt{node\_a} when \texttt{node\_b} was next \\[2pt]
\textsc{terminate}
  & Hard-halt the process
  & Process instance is closed; no further nodes execute \\[2pt]
\textsc{remove\_node}
  & Remove a node from the topology
  & Engine rewires the node's predecessor and successor flows to bypass it
    and resumes at the correct point \\[2pt]
\textsc{add\_node}
  & Insert a new task node
  & Engine wires flows through the new node before the current target and
    resumes at the inserted node \\
\bottomrule
\end{tabularx}
\caption{Intervention primitives available to the FlowAgent and their structural
effects on the process topology.}
\label{tab:actions}
\end{table}

Any of the three agent types may employ a designated \texttt{ask\_user} tool to solicit user input directly, either via a local interaction channel or, when the MCP bridge supports it, delegated to the workflow engine.

\medskip
The \textsc{skip\_to} action is particularly powerful: equipped with the
remaining-tasks dictionary, the hook LLM can jump to any node in the graph,
effectively reshuffling the remaining process. A hook at the midpoint of a
workflow can skip ahead to the final task or bypass multiple intermediate steps
based on a compliance condition. The first five actions are runtime navigation
decisions over the existing node set. \textsc{remove\_node} and \textsc{add\_node}
go further: they trigger a \emph{topology modification} and may only target nodes
that have not yet executed. The workflow engine applies the structural change and resumes
execution at the correct entry point without replaying already-executed nodes.

\paragraph{Separation of concerns: instruct vs.\ execute.}
CUGA FLO issues the hook action \emph{instruction}. The engine carries it out.
Executing the action is the responsibility of the WorkflowEngine, which receives
the \texttt{HookResult} via the MCP bridge and applies the corresponding structural
intervention (routing, graph rebuild, halt) according to its own execution model.
Per-process \texttt{action\_permissions}, declared in the process configuration,
explicitly list which hook actions are permitted or prohibited for a given process,
providing an additional governance layer over what adaptations the FlowAgent may
instruct. Hook reasoning is performed by the FlowAgent itself (not a separate
agent) because hooks are a process-level concern, and every deviation is
policy-governed and recorded in the process audit log.

\subsection{The DecisionAgent: Policy-Governed Routing}

Each decision gateway (i.e., conditional split) that requires agent-based routing
is bound to a \texttt{DecisionAgent}. The DecisionAgent is a specialized \texttt{CugaAgent} wrapper
that implements two-step routing.

In \textbf{Step~1}, the condition expression associated with the gateway is taken as
input. Each \texttt{\$\{var\}} token is bound to its current value from the process
variables, and the resulting expression is evaluated deterministically to identify
the subset of outbound flows to be traversed.

In \textbf{Step~2}, the result of the condition evaluation, together with the full
process knowledge $P$ and the gateway's decision policy $C_D$, is introduced to
the decision agent. The agent reasons over this context to conclude which flows should be
traversed next, returning the chosen flow ID(s) validated against the known outgoing
flows.

\subsection{The TaskAgent: Policy-Constrained Workload Execution}

TaskAgents wrap around \texttt{CugaAgent} instances bound to individual process tasks.
This is the default selection in CUGA FLO. Per the application configuration YAML
(see Section~\ref{sec:deployment-yamls-schemas}), the \texttt{agent\_type} field
allows other agent frameworks to be employed instead.
Each TaskAgent receives, at invocation time, a composite input constructed by
the FlowAgent from three sources:

\begin{enumerate}[noitemsep, label=(\roman*)]
  \item The task goal statement.
  \item The task policy $C_T(t_i)$ (the governing FRAME policy).
  \item The \texttt{ControlPointFlowKnowledge}, which embeds the current process
        variables, task results from preceding tasks, and all contextual information
        about the process state, history, and model required to fulfill the task.
\end{enumerate}

The task policy $C_T$ frames the behavioral constraints within which the TaskAgent
pursues its goal. Rather than specifying the goal itself, $C_T$ defines what the
agent may and may not do, what outputs it is expected to produce, and any
domain-specific restrictions it must observe, for instance the credit check task
must not render an approval decision, as that responsibility belongs to the gateway.
The agent is driven by its goal statement; $C_T$ bounds the space of admissible
behavior.

A TaskAgent may use any tools available in the CUGA tool registry or MCP-connected
servers. It assembles these tools autonomously under its \texttt{CugaAgent} execution
loop, returning a natural-language result. The TaskAgent communicates its output to
the workflow engine.

\subsection{ControlPointFlowKnowledge: Observable Shared Process Knowledge}

\begin{sloppypar}
\texttt{ControlPointFlowKnowledge} is the information structure the workflow engine
sends to CUGA FLO at every control point callback. It combines control-point
identity fields (\texttt{element\_id}, \texttt{element\_name}) with
control-point-specific context (\texttt{task\_instruction} for task nodes,
\texttt{available\_flows} for gateways, and \texttt{edge\_id} for hooks) and an
embedded \texttt{FlowState} snapshot (\texttt{current\_state}) that carries the
complete process knowledge required for agent reasoning. Key fields of the embedded state available for agent reasoning include:
\texttt{process\_variables} (the shared data store updated by task agents and
consulted by decision agents and the FlowAgent), \texttt{execution\_path} (the
ordered sequence of activated nodes, forming an audit trail), \texttt{task\_results}
(per-task outputs), and \texttt{gateway\_decisions} (routing decisions audit).
The embedded state also carries terminal outcome fields
(\texttt{is\_complete}, \texttt{is\_halted}, \texttt{halt\_reason}), which are
written by the workflow engine to signal process execution outcome.
Process variables constitute the observable substrate of process awareness: task
agents set them, decision agents read them for routing, and the FlowAgent may read
and override them at hook points. All coordination between agents flows through this
observable, serializable structure, with no direct inter-agent communication.
\end{sloppypar}

\subsection{The Integration Layer: MCP Bridge and Engine}
\label{sec:integration}

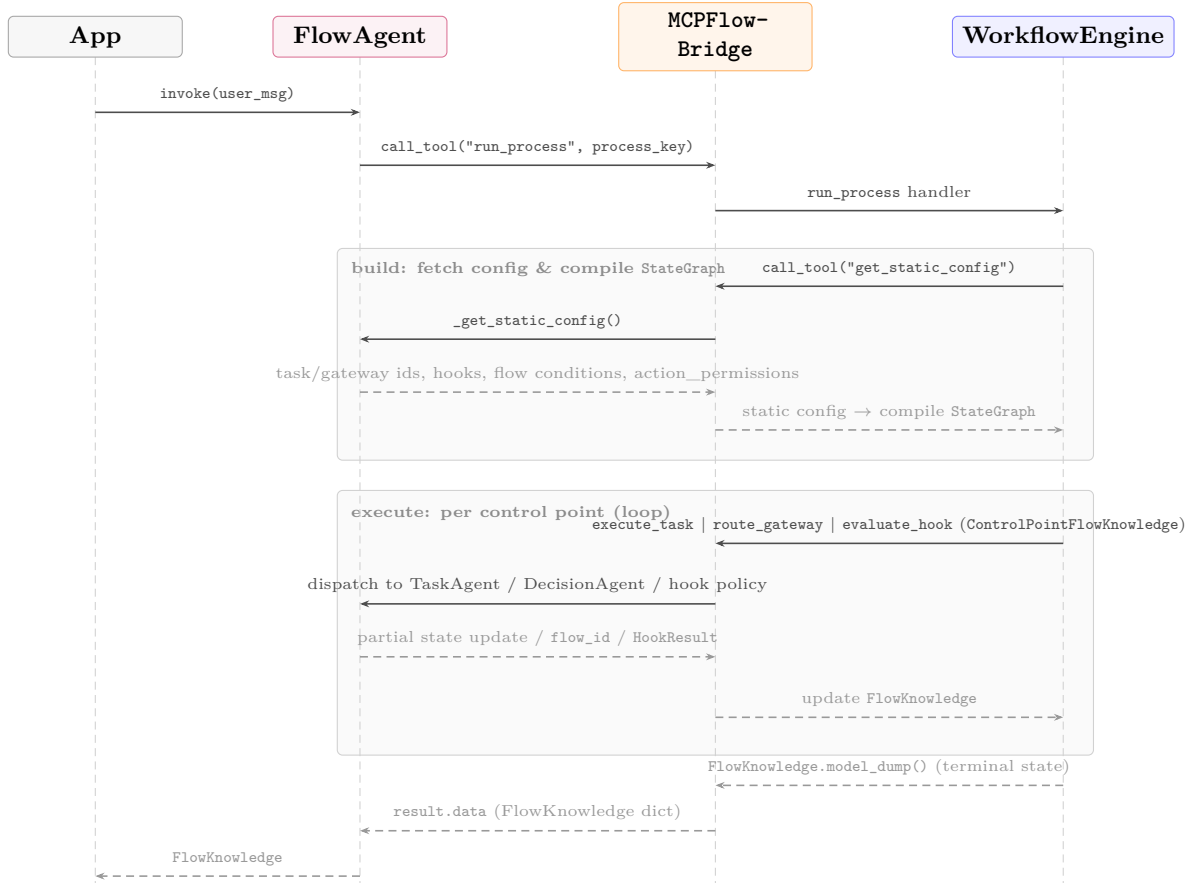
\begin{figure}[H]
\centering
\begin{tikzpicture}[
  seqact/.style={rectangle, rounded corners=2pt, draw=black!50,
                 minimum width=2.3cm, minimum height=0.5cm,
                 text centered, font=\footnotesize\bfseries, fill=white},
  msg/.style={-{Stealth[length=3.5pt]}, semithick, black!70},
  ret/.style={-{Stealth[length=3.5pt]}, semithick, black!40, densely dashed},
]

\def\xA{0}     
\def\xF{3.5}   
\def\xB{8.2}   
\def\xE{12.8}  

\begin{scope}[on background layer]
  \draw[rounded corners=2pt, draw=black!18, fill=gray!4]
       (\xF-0.3,-2.8) rectangle (\xE+0.4,-5.6);
  \draw[rounded corners=2pt, draw=black!18, fill=gray!4]
       (\xF-0.3,-6.0) rectangle (\xE+0.4,-9.5);
\end{scope}

\node[font=\tiny\bfseries\color{black!45}, anchor=north west]
     at (\xF-0.25,-2.85) {build: fetch config \& compile \texttt{StateGraph}};
\node[font=\tiny\bfseries\color{black!45}, anchor=north west]
     at (\xF-0.25,-6.05) {execute: per control point (loop)};

\draw[black!18, densely dashed, thin] (\xA,-0.27) -- (\xA,-11.2);
\draw[black!18, densely dashed, thin] (\xF,-0.27) -- (\xF,-11.2);
\draw[black!18, densely dashed, thin] (\xB,-0.27) -- (\xB,-11.2);
\draw[black!18, densely dashed, thin] (\xE,-0.27) -- (\xE,-11.2);

\node[seqact, draw=black!35,   fill=gray!6]    at (\xA,0) {App};
\node[seqact, draw=purple!55,  fill=purple!5]  at (\xF,0) {FlowAgent};
\node[seqact, draw=orange!60,  fill=orange!8,  text width=2.3cm]
     at (\xB,0) {\texttt{MCPFlow-}\\\texttt{Bridge}};
\node[seqact, draw=blue!50,    fill=blue!6]    at (\xE,0) {WorkflowEngine};

\draw[msg] (\xA,-1.0) -- (\xF,-1.0)
     node[midway,above,font=\tiny]{\texttt{invoke(user\_msg)}};

\draw[msg] (\xF,-1.7) -- (\xB,-1.7)
     node[midway,above,font=\tiny]{\texttt{call\_tool("run\_process", process\_key)}};
\draw[msg] (\xB,-2.3) -- (\xE,-2.3)
     node[midway,above,font=\tiny]{\texttt{run\_process} handler};

\draw[msg] (\xE,-3.3) -- (\xB,-3.3)
     node[midway,above,font=\tiny]{\texttt{call\_tool("get\_static\_config")}};
\draw[msg] (\xB,-4.0) -- (\xF,-4.0)
     node[midway,above,font=\tiny]{\texttt{\_get\_static\_config()}};
\draw[ret] (\xF,-4.7) -- (\xB,-4.7)
     node[midway,above,font=\tiny]{task/gateway ids, hooks, flow conditions, action\_permissions};
\draw[ret] (\xB,-5.2) -- (\xE,-5.2)
     node[midway,above,font=\tiny]{static config $\rightarrow$ compile \texttt{StateGraph}};

\draw[msg] (\xE,-6.7) -- (\xB,-6.7)
     node[midway,above,font=\tiny]{\texttt{execute\_task} $|$ \texttt{route\_gateway} $|$ \texttt{evaluate\_hook}
     (\texttt{ControlPointFlowKnowledge})};
\draw[msg] (\xB,-7.5) -- (\xF,-7.5)
     node[midway,above,font=\tiny]{dispatch to TaskAgent / DecisionAgent / hook policy};
\draw[ret] (\xF,-8.2) -- (\xB,-8.2)
     node[midway,above,font=\tiny]{partial state update / \texttt{flow\_id} / \texttt{HookResult}};
\draw[ret] (\xB,-9.0) -- (\xE,-9.0)
     node[midway,above,font=\tiny]{update \texttt{FlowKnowledge}};

\draw[ret] (\xE,-9.9) -- (\xB,-9.9)
     node[midway,above,font=\tiny]{\texttt{FlowKnowledge.model\_dump()} (terminal state)};
\draw[ret] (\xB,-10.5) -- (\xF,-10.5)
     node[midway,above,font=\tiny]{\texttt{result.data} (FlowKnowledge dict)};
\draw[ret] (\xF,-11.1) -- (\xA,-11.1)
     node[midway,above,font=\tiny]{\texttt{FlowKnowledge}};

\end{tikzpicture}
\caption{CUGA FLO runtime interaction sequence. Solid arrows are calls; dashed
arrows are returns. \texttt{MCPFlowBridge} is the sole channel between the process
process harness (\texttt{FlowAgent}) and the execution layer (\texttt{WorkflowEngine}).
The \emph{build} phase (once per invocation) fetches static config so the engine can
compile its execution graph. The \emph{execute} phase issues one
round-trip per control point (task nodes, gateways, and hook intercept
flows), passing a \texttt{ControlPointFlowKnowledge} and receiving structured results
that the engine uses to update its process knowledge. No direct reference between
the process harness and \texttt{WorkflowEngine} exists. The transport interface decouples CUGA FLO from the workflow engine, allowing the workflow engine implementation to be replaced transparently.}
\label{fig:sequence}
\end{figure}

The defining architectural choice of CUGA FLO is that the process harness and the
execution layer (the workflow engine) are fully decoupled and communicate only
through a Model Context Protocol (MCP) bridge\footnote{\url{https://modelcontextprotocol.io}}. This makes the execution
backend replaceable without any change to the reasoning layer.
The runtime interaction sequence is shown in Figure~\ref{fig:sequence}.

\paragraph{MCPFlowBridge.}
The \texttt{MCPFlowBridge} is a FastMCP server that acts as the integration contract
between the process harness and any \texttt{WorkflowEngine}. The FlowAgent side
registers four callback listeners as MCP tools: \texttt{execute\_task} (a task
node is reached), \texttt{route\_gateway} (a gateway needs a routing decision),
\texttt{evaluate\_hook} (a hook intercept point fires), and
\texttt{get\_static\_config} (the engine fetches process metadata at startup). The
engine side registers \texttt{run\_process} (start a new process instance). Every
call carries a \texttt{ControlPointFlowKnowledge}, so each reasoning call is
self-contained and the engine is stateless from the FlowAgent's perspective.

\paragraph{WorkflowEngine.}
CUGA FLO defines an abstract \texttt{WorkflowEngine} interface as the integration
API required to connect any workflow engine to the process harness. It holds the
process model and instance state, drives execution node by node, and communicates
with the FlowAgent exclusively through the bridge, calling the corresponding
callback at each control point. The implementation includes a
\texttt{LangGraphWorkflowEngine}, a minimal instantiation that compiles the BPMN
topology into a LangGraph \texttt{StateGraph}, intended to be replaced by an
enterprise-grade workflow engine in a production environment. This is the sense in
which the structural conformance guarantee of CUGA FLO is engine-enforced: the \texttt{LangGraphWorkflowEngine} executes the actual process topology, and the FlowAgent governs the
policy-bounded reasoning at designated control points.

Figure~\ref{fig:sequence} shows the runtime interaction sequence.

\subsection{Application Deployment}
\label{sec:deployment-yamls-schemas}

A CUGA FLO application is deployed as a self-contained directory organized into two
subfolders: \texttt{config/} and \texttt{policies/}.

\paragraph{The \texttt{config/} folder.}
The \texttt{config/} folder holds the process definition and the two YAML
configuration files that govern the deployment.

\begin{itemize}[noitemsep]
  \item \textbf{Process model file} (\texttt{.bpmn}): the workflow process
        definition, referenced by the application YAML and compiled by the
        \texttt{WorkflowEngine} at startup into an executable process graph.

  \item \textbf{Application YAML} (\texttt{<app>\_config.yaml}): the primary
        configuration artifact, conforming to the \texttt{AppYaml} Pydantic schema.
        Its top-level blocks are:
        \begin{itemize}[noitemsep]
          \item \texttt{flow:} — process identity: name, identifier, version,
                path to the process model file, and the agent framework employed
                by the FlowAgent (\texttt{agent\_type}, default \texttt{cuga\_agent}).
          \item \texttt{variables:} — initial values of process variables shared
                across all agent types during execution.
          \item \texttt{tasks:} — one entry per process task, declaring its
                \texttt{mode} (\texttt{task\_agent} or \texttt{native}) and, for
                \texttt{task\_agent} tasks, the agent name, system instruction, tools,
                path to the task policy document, and optionally the agent framework
                for this task (\texttt{agent\_type}, default \texttt{cuga\_agent}).
          \item \texttt{gateways:} — one entry per conditional gateway, declaring its
                \texttt{mode} (\texttt{decision\_agent} or \texttt{native}), the
                condition expression for deterministic pre-evaluation, the path to the
                decision policy document, optional per-flow decision labels, and
                optionally the agent framework for the decision agent
                (\texttt{agent\_type}, default \texttt{cuga\_agent}).
          \item \texttt{action\_permissions:} — per-process lists of hook actions
                explicitly permitted or prohibited, bounding the structural
                interventions the FlowAgent may instruct for this process.
          \item \texttt{hooks:} — one entry per hook, specifying its identifier,
                the flow identifier it intercepts (\texttt{location}), and the path to the hook
                policy document.
          \item \texttt{llm:} — optional override of the default LLM provider, model,
                and temperature for all agents in this process.
        \end{itemize}

  \item \textbf{Supervisor YAML} (\texttt{supervisor\_<app>.yaml}): configures the
        \texttt{CugaSupervisor} that presents a conversational interface over the
        process, conforming to the \texttt{SupervisorYaml} Pydantic schema. It
        declares the supervisor model, description, and special instructions, and
        lists the \texttt{FlowAgent} instances it coordinates, each identified by
        name, type, and a reference to its application YAML.
\end{itemize}

\paragraph{The \texttt{policies/} folder.}
The \texttt{policies/} folder contains the markdown policy documents that constitute
the process FRAME. Each file is referenced by path from the application YAML and
loaded at \texttt{FlowAgent} initialization. By convention, file names encode the
agent type and the process element they govern: \texttt{task-<name>.md} for task
policies, \texttt{decision-<name>.md} for gateway decision policies, and
\texttt{hook-<name>.md} for hook policies. Being plain markdown files, policies are
human-readable, version-controllable, and updatable independently of the process
model or the application code, with changes taking effect on all subsequent
reasoning calls within running instances.

\section{Case Study: Loan Approval Workflow}
\label{sec:casestudy}

\subsection{Process Overview}

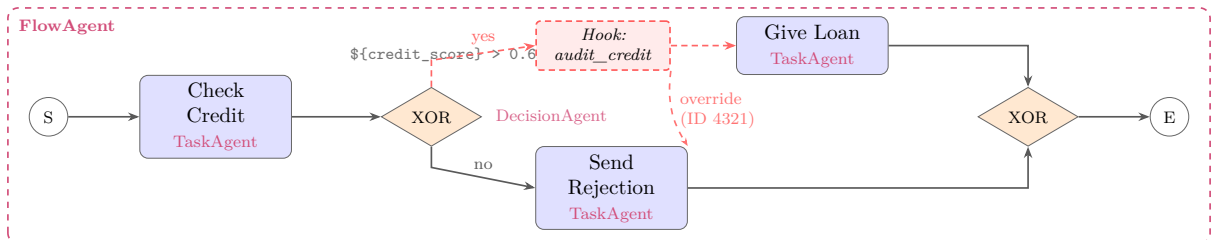
\begin{figure}[H]
\centering
\resizebox{\linewidth}{!}{\begin{tikzpicture}[node distance=0.7cm and 1.4cm]

\node[event] (start) {S};

\node[task, right=1.2cm of start, yshift=0pt] (credit)
      {Check\\Credit\\{\scriptsize\color{purple!70}TaskAgent}};

\node[gateway, right=1.5cm of credit] (gw1) {XOR};
\node[font=\scriptsize, above=0.3cm of gw1, xshift=0.2cm, color=black!55]
      {\texttt{\$\{credit\_score\}~>~0.6}};
\node[font=\scriptsize, right=0.1cm of gw1, color=purple!60]
      {DecisionAgent};

\node[hook, right=0.9cm of gw1, yshift=1.2cm] (hook)
      {Hook:\\audit\_credit};

\node[task, right=1.1cm of hook] (loan)
      {Give Loan\\{\scriptsize\color{purple!70}TaskAgent}};

\node[task, right=0.9cm of gw1, yshift=-1.2cm] (reject)
      {Send\\Rejection\\{\scriptsize\color{purple!70}TaskAgent}};

\node[gateway, right=1.5cm of loan, yshift=-1.2cm] (gw2) {XOR};

\node[event, right=1.2cm of gw2] (end) {E};

\draw[flow] (start) -- (credit);
\draw[flow] (credit) -- (gw1);
\draw[hookflow] (gw1.north) -- ++(0,0.45) -- node[above,font=\scriptsize]{yes} (hook.west);
\draw[hookflow] (hook) -- (loan);
\draw[flow] (gw1.south) -- ++(0,-0.35) -- node[above,font=\scriptsize]{no} (reject.west);
\draw[flow] (loan.east) -| (gw2.north);
\draw[flow] (reject.east) -| (gw2.south);
\draw[flow] (gw2) -- (end);

\draw[-{Stealth[length=4pt]}, red!55, thick, dashed, bend right=20]
      (hook.south east) to node[right, align=left, font=\scriptsize\color{red!60}]
      {override\\(ID 4321)} (reject.north east);

\begin{scope}[on background layer]
  \draw[dashed, purple!70, rounded corners=5pt, thick]
    ([xshift=-0.45cm, yshift=-1.9cm]start.south west)
    rectangle
    ([xshift=0.45cm, yshift=1.6cm]end.north east);
\end{scope}
\node[font=\scriptsize\bfseries\color{purple!70}, anchor=north west]
  at ([xshift=-0.4cm, yshift=1.55cm]start.north west) {FlowAgent};

\end{tikzpicture}}
\caption{Loan approval BPMN model with TDF annotations. The red dashed hook
node intercepts the ``yes'' sequence flow; the curved
red arrow shows the \textsc{skip\_to} redirect for applicant ID~4321.}
\label{fig:bpmn}
\end{figure}

We demonstrate CUGA FLO through a loan approval workflow that instantiates all
three TDF agent types. The workflow, illustrated in~\Cref{fig:bpmn}, evaluates personal loan applications in four
stages: credit assessment, approval routing, outcome handling, and merge.

\subsubsection{BPMN Structure}

\subsection{FRAME Instantiation}

The FRAME $\mathcal{F}$ for this process comprises four policy documents:

\subsubsection{Task Policy: Credit Check}

\begin{policybox}
\textbf{Credit Check Policy (abbreviated).}
Analyse the applicant's financial profile and produce a normalised
\texttt{credit\_score} $\in [0,1]$. Score using: payment history (35\%),
debt-to-income (30\%), history length (15\%), credit mix (10\%), recent
enquiries (10\%). Set \texttt{credit\_score} before completing. \emph{Do not
approve or reject. That decision belongs to the gateway.}
\end{policybox}

The explicit prohibition ``do not approve or reject'' is the policy-level
enforcement of TDF separation: the task agent is constrained from performing
gateway routing, even though it has the credit score value.

\subsubsection{Decision Policy: Credit Gateway}

\begin{policybox}
\textbf{Decision Policy: Gateway\_09ad5fc.}
Route using \texttt{\${credit\_score} > 0.6}: scores above 0.60 to the
approval path (\texttt{Flow\_0ybszcv}); at or below to rejection
(\texttt{Flow\_1jgea85}). Override: if \texttt{loan\_amount > 50{,}000} apply
strict threshold 0.75. Output exactly one flow ID.
\end{policybox}

For the typical case (\texttt{credit\_score} is a float), Step~1 resolves this
deterministically. The LLM is called only when \texttt{credit\_score} is absent
or the higher threshold override applies.

\subsubsection{Hook Policy: Regulatory Override}

\begin{policybox}
\textbf{Hook Policy: audit\_credit\_decision.}
\textbf{Rule}: Applicant ID \texttt{4321} is subject to regulatory restriction;
redirect to rejection. Set \texttt{rejection\_reason =}\\
\texttt{"regulatory\_policy\_restriction"}\\
and \texttt{policy\_override\_active = true}.
\textbf{Default}: All other IDs: return \textsc{continue} with no state
updates.
\end{policybox}

\subsubsection{Task Policies: Outcome Handling}

The loan processor policy requires verifying \texttt{approved == true} before
disbursing and setting \texttt{loan\_granted = true}; it must not send rejections.
The rejection handler policy routes between two paths based on
\texttt{rejection\_reason}: a standard credit-score rejection, or a regulatory
override path that cites regulatory compliance without referencing the credit score.

\subsection{Process Application Deployment}

The loan approval application follows the two-folder deployment structure described
in Section~\ref{sec:deployment-yamls-schemas}.

\paragraph{\texttt{config/}.}
The \texttt{config/} folder contains three files.
\texttt{BPMNdiagram.bpmn} holds the process model with five elements: three task
nodes (\emph{Check Credit}, \emph{Give Loan}, \emph{Send Rejection}), one
conditional XOR split gateway, and one merge gateway.
\texttt{loan\_approval\_config.yaml} declares three \texttt{task\_agent} tasks (one
per task node), one \texttt{decision\_agent} gateway (the credit split, with
condition \texttt{\$\{credit\_score\} > 0.6} and a reference to the decision policy
document), one \texttt{native} gateway (the merge), and one hook
(\texttt{audit\_credit\_decision}, type \texttt{edge}, intercepting flow
\texttt{Flow\_0ybszcv}).
Action permissions for this process permit \textsc{continue}, \textsc{skip\_to},
and \textsc{terminate}, while prohibiting \textsc{swap\_nodes} and \textsc{skip\_node}.
\texttt{supervisor\_loan\_approval.yaml} configures the \texttt{CugaSupervisor} with
a single registered \texttt{FlowAgent} (\texttt{loan\_flow\_agent}) pointing to the
application YAML, and provides special instructions for presenting approval and
regulatory override outcomes to the end user.

\begin{sloppypar}
\paragraph{\texttt{policies/}.}
The \texttt{policies/} folder contains five markdown documents forming the FRAME
$\mathcal{F}$ for this process:
\begin{itemize}[noitemsep]
  \item \texttt{task-credit\_check.md}: constrains the credit checker to produce a
        normalised \texttt{credit\_score} $\in [0,1]$ without rendering an approval
        or rejection decision.
  \item \texttt{decision-credit\_decision.md}: governs routing at the credit gateway,
        specifying the primary threshold and a stricter override for high loan amounts.
  \item \texttt{hook-audit\_credit\_decision.md}: the regulatory override policy,
        redirecting applicant ID~4321 to the rejection path via \textsc{skip\_to}.
  \item \texttt{task-loan\_processor.md}: constrains the loan processor to verify
        approval status and record the grant without sending rejections.
  \item \texttt{task-rejection\_handler.md}: governs the rejection handler,
        distinguishing standard credit-score rejections from regulatory override paths.
\end{itemize}
\end{sloppypar}

\subsection{FlowAgent Instantiation}

In the application deployment, the \texttt{FlowAgent} is instantiated by passing a
\texttt{process\_key} that identifies the application YAML registered on the
\texttt{MCPFlowBridge} as follows:

\begin{lstlisting}[style=python]
loan_flow_agent = FlowAgent(process_key="loan_approval_v1")
\end{lstlisting}

At construction time, \texttt{FlowAgent} calls
\texttt{bridge.get\_flow\_annotations(process\_key)}, which returns a
\texttt{FlowConfig} parsed from \texttt{loan\_approval\_config.yaml}.
\texttt{FlowConfig} then constructs all durable state via its factory methods:
three \texttt{TaskAgent} instances (one per \texttt{task\_agent} task entry),
one \texttt{DecisionAgent} (for the \texttt{decision\_agent} gateway, loading
its condition and policy document), the hook list (one \texttt{Hook} for
\texttt{audit\_credit\_decision}), the action permissions, and the initial
process variable values. The merge gateway, declared \texttt{native}, requires no
agent and is routed inline by the \texttt{FlowAgent} via condition evaluation.
No process model, execution state, or graph is held by the \texttt{FlowAgent};
these remain the responsibility of the \texttt{WorkflowEngine}.

\subsection{Execution Traces}

\paragraph{Nominal case (ID~23, score~0.6625).}
\begin{enumerate}[noitemsep, label=\arabic*.]
  \item TaskAgent: credit check → \texttt{credit\_score = 0.6625}
  \item DecisionAgent: Step-1 eval of \texttt{0.6625 > 0.6} → \texttt{TRUE}
        → record \texttt{Flow\_0ybszcv}
  \item Hook (ID~23): LLM evaluates policy → \textsc{continue} → proceed to
        loan processing
  \item TaskAgent: loan processor → \texttt{loan\_granted = true}
\end{enumerate}

\paragraph{Regulatory block (ID~4321, score~0.82).}
\begin{enumerate}[noitemsep, label=\arabic*.]
  \item TaskAgent: credit check → \texttt{credit\_score = 0.82}
  \item DecisionAgent: Step-1 eval → \texttt{TRUE} → record \texttt{Flow\_0ybszcv}
  \item Hook (ID~4321): LLM evaluates policy → \textsc{skip\_to}
        \texttt{Activity\_131ar38};\\
        patches \texttt{rejection\_reason =}
        \texttt{"regulatory\_policy\_restriction"}
  \item TaskAgent: rejection handler → regulatory path (no credit score reference)
\end{enumerate}

This trace illustrates the core Agentic BPM property: the DecisionAgent routes
correctly by credit score (0.82 \textgreater\ 0.6), while the hook overrides
the resulting edge based on a wholly different criterion (applicant ID), without
modifying the gateway policy. The two concerns are independently governed by
separate FRAME policies, each auditable in isolation. The hook is attached exclusively to \texttt{Flow\_0ybszcv} (the
``yes'' arc), so it fires only when the credit decision would result in approval;
the rejection path bypasses the hook entirely.

\section{Related Work}
\label{sec:related}

\paragraph{From AI-augmented to Agentic BPM.}
The trajectory of research leading to CUGA FLO follows a clear paradigm
progression. Dumas et al.~\cite{dumas2023aiaugmented} introduced the AI-augmented
BPM (ABPMS) vision, in which AI capabilities are embedded into BPM systems to
make processes more adaptive, proactive, and context-sensitive. This laid the
foundation for the \emph{Agentic BPM} paradigm, formally articulated as a research
manifesto in~\cite{calvanese2026agenticbpm}: agents act as primary process-aware functional
entities that perceive, reason, and act under framed autonomy, with explainability, and conversational actionability as core capabilities. This manifesto has been a primary driver leading to the development of the process harness concept, with CUGA FLO as its first fully realized implementation. Further to the Agentic BPM paradigm articulated in the recent manifesto, Dumas et al.~\cite{dumas2026abpms} elaborate on the vision of Agentic BPM systems and their evolution along a continuum of autonomy, while outlining a reference architecture for their realization. In line with this vision, CUGA FLO provides a pragmatic execution and governance framework for realizing this transition, enabling organizations to incrementally augment workflow-driven processes with agentic tasks, decisions, and flow interventions while preserving the structural guarantees of the underlying process engine.

\paragraph{LLM agents, planning, and workflow automation.}
Contemporary agentic LLM frameworks are typically built around an inner execution
loop that iteratively alternates between LLM prompt preparation, LLM inference, and
tool invocations. ReAct~\cite{yao2022react}, for example, interleaves reasoning and
tool use without enforcing prior condition evaluation. The DecisionAgent addresses
this by mandating a deterministic condition-evaluation step before any LLM-based
reasoning, ensuring the gateway condition is explicitly assessed before policy-aware
routing proceeds. AutoGen~\cite{wu2023autogen} and CrewAI~\footnote{\url{https://github.com/crewaiinc/crewai}} organize
role-specialized agents, whereas TDF provides a formal semantic layer for such roles.
LLM-based planning~\cite{kambhampati2024llm} identifies plan correctness as an open
problem that TDF's structural conformance addresses. RPA executes scripted bots within
BPMN structures (i.e., Augmented BPM), whereas CUGA FLO replaces them with LLM agents
while preserving structural guarantees (i.e., Agentic BPM). Conversational workflow
systems~\cite{qian2024chatdev} coordinate agents through dialogue, whereas TDF enforces
ordering guarantees through an underlying workflow engine that executes the process
topology deterministically, while preserving conversational actionability through
agent-accessible user interaction via the \texttt{ask\_user} tool.

\paragraph{Policy, governance, and process-aware AI.}
Constitutional AI~\cite{bai2022constitutional} governs individual agent behavior through normative principles, whereas CUGA FLO embeds governance directly into workflow execution via the FRAME, enforcing policies over tasks, decisions, and flow interventions.
Intelligent process
automation~\cite{geyer2021intelligent} uses ML for decision support, whereas CUGA FLO
makes the agent the process overseer and executor. Adaptive case
management~\cite{swenson2010mastering} addresses flexible processes through rule-based
mechanisms, whereas Agentic BPM replaces rule-based adaptation with LLM reasoning
bounded by the FRAME.

\paragraph{Agents in BPM.}
Prior work has incorporated agents into BPM for specific purposes, including
agent-assisted task execution, process monitoring, and intelligent resource
allocation within conventional BPM pipelines~\cite{dumas2023aiaugmented,calvanese2026agenticbpm}. These approaches treat agents as
point augmentations of specific activities within an otherwise conventional BPM
system, leaving the process structure and execution model unchanged. CUGA FLO is,
to our knowledge, the first to propose a complete model, the TDF
process harness, for systematically transforming any workflow system into an agentic
one, with principled separation across task execution, routing, and flow
supervision, and a formal account of how process awareness and policy framing are
instantiated at each agent type.

\subsection{Comparison with Related Paradigms}
\label{sec:comparison}

\subsubsection{Property Analysis}

We compare CUGA FLO against reference approaches along key properties (see~\Cref{tab:comparison}).

\begin{table}[H]
\centering
\resizebox{\textwidth}{!}{%
\begin{tabular}{@{}lp{1.6cm}p{1.6cm}p{1.6cm}p{1.8cm}p{1.8cm}@{}}
\toprule
\textbf{Property} &
\centering\textbf{Classical BPM} &
\centering\textbf{LLM-as-Planner} &
\centering\textbf{ReAct/ AutoGen} &
\centering\textbf{Agentic Workflows} &
\centering\arraybackslash\textbf{CUGA FLO (TDF)} \\
\midrule
Process graph enforced & \checkmark & Input only & \texttimes & Dev-defined & \checkmark\ (executed) \\
Structural conformance & \checkmark & \texttimes & \texttimes & Dev-defined & \checkmark \\
Runtime adaptability & \texttimes & \checkmark & \checkmark & \checkmark & \checkmark \\
Human-readable policies & \texttimes & \texttimes & \texttimes & \texttimes & \checkmark \\
Policy accountability & \texttimes & \texttimes & Implicit & Implicit & \checkmark \\
Audit trail (per element) & \checkmark & \texttimes & \texttimes & \texttimes & \checkmark \\
Exception: design-time & \checkmark & \texttimes & \texttimes & \texttimes & Optional \\
Exception: open-world & \texttimes & \checkmark & \checkmark & \checkmark & \checkmark \\
Repeatable traces & \checkmark & \texttimes & \texttimes & \texttimes & \checkmark \\
Non-programmer authoring & \texttimes & \texttimes & \texttimes & Partial & \checkmark \\
\bottomrule
\end{tabular}%
}
\caption{Property comparison across process execution paradigms. Agentic Workflows
denotes dynamic, LLM-devised execution plans per a given intent (e.g., orchestrated
sub-agent pipelines), which achieve open-world adaptability but without integration
with a structured, enterprise-grade workflow engine and without structural
conformance guarantees.}
\label{tab:comparison}
\end{table}

\subsubsection{The Key Distinctions}

\paragraph{vs.\ Classical BPM exception handling.}
Classical BPM exception handling is \emph{additive} and \emph{design-time}: every
deviation path must be pre-modeled as a boundary event, error sub-process, or
compensation handler. CUGA FLO hooks are \emph{open-world}: the set of situations
a hook policy can handle is the set of situations the policy document can reason
about. In the loan example, the regulatory rule is a markdown policy reasoned about
per case, not a diagram element. Adding a new exception in classical BPM requires
a model change (release-gated). In CUGA FLO it is a policy document update that
may be applied at any time, taking effect on all subsequent reasoning calls within
running instances.

\paragraph{vs.\ LLM-as-Planner.}
LLM-as-planner treats the process model as \emph{input}: the LLM reads it and
constructs an ad~hoc plan without enforcement. CUGA FLO is fundamentally different:
the workflow engine executes the process topology directly, making non-conforming
execution physically impossible. Adaptations happen at runtime with actual process
state (not predicted context), and every activated node, gateway decision, and hook
evaluation is recorded in the process knowledge tied to process element identifiers.
Two planning-LLM invocations with identical inputs may produce structurally different
traces. CUGA FLO traces are structurally determined by the process topology, deterministically executed by the workflow engine.

\paragraph{vs.\ Agentic Workflows.}
Agentic workflow systems, such as ad-hoc devised and orchestrated sub-agent
sub-stepped pipelines (also known as chain-of-thought
prompting~\cite{Wei2022CoT}) that dynamically devise an execution plan from a
high-level intent, as in dynamic workflows realized in systems such as Claude
Code\footnote{\url{https://claude.com/blog/introducing-dynamic-workflows-in-claude-code}},
differ from LLM-as-planner
approaches primarily in execution discipline: a coordinator agent decomposes a goal
into steps and delegates each to a specialized sub-agent, collecting and chaining
their results. However, the core distinction from CUGA FLO remains: the execution
plan is assembled dynamically by the coordinating LLM and is not bound to a
pre-specified, engine-enforced process topology. Consequently, there is no
structural conformance guarantee, no per-element audit trail tied to a process
model, and no formal policy accountability at the task, routing, or flow level.
CUGA FLO differs fundamentally: the process topology is fixed and executed
deterministically by the workflow engine, and the process harness augments it at designated
control points under explicit policies, rather than replacing it with a dynamically
constructed plan. This distinction is especially significant in enterprise settings,
where cross-cutting workflows must satisfy regulatory, audit, and compliance
requirements that ad~hoc orchestration cannot guarantee.

\section{Discussion and Conclusion}
\label{sec:discussion}

\subsection{Novelty of the Process Harness Mechanism}

This work introduces a new form of process instrumentation: the \textbf{process
harness}. We formally characterize the process harness through the TDF model, specifying
its data schema and execution semantics, and provide a concrete realization in CUGA
FLO. The instrumentation uplifts any existing workflow system into a fully agentic
one by integrating policy-governed LLM reasoning with deterministic process
execution, while remaining agnostic to the internal workings of the underlying
workflow engine and to the specific agentic resources it employs.

The TDF model is, to our knowledge, the first to embed LLM agents as
policy-bounded \emph{governing authorities} over a workflow execution graph rather
than as consumers of a process description. Classical BPM robustness is attained either through rigid situational handling that
must be pre-built into the model at design time, or through escalation to external
human users. The process harness is interceptive by design: it integrates agentic
reasoning at designated control points for coherent, policy-governed situational
resolution, reducing dependence on human escalation. Navigation actions select among valid nodes in the
process graph, while topology actions modify the graph
under policy and are then executed by the engine, so the mechanism is genuinely
open-world while remaining structurally sound; it is also auditable because every hook invocation and its resulting intervention are recorded in the process knowledge, each tied to its governing policy document.

The realization of the FRAME $\mathcal{F}$ by associating each agent reasoning loop
with a policy provides an accountability substrate absent from LLM-as-planner
approaches: every LLM call can be audited against its governing policy document, and
non-compliance is diagnosable without inspecting code.

\subsection{Limitations and Future Work}

\paragraph{Process model coverage and topology modification.}
Intermediate events can be treated as hook attachment points in the same way
sequence flows are annotated. This is planned as an explicit first-class construct
in a future release. Composite tasks (comprising internal sub-processes) can be
treated as a separate flow agent handling all sub-process callbacks from the
perspective of a parent flow agent, yielding a
hierarchical structure of nested flow agents, each governing its own sub-process
independently. Runtime topology modification (inserting or removing a task node
relative to the original process model) is supported through the
\textsc{add\_node} and \textsc{remove\_node} hook actions: the engine modifies the
live process model, recompiles, and resumes at the correct entry point without
replaying already-executed nodes. In the current implementation, topology modifications (\textsc{add\_node} and \textsc{remove\_node}) may only target nodes that have not yet executed; applying them to nodes that are already active or completed is not supported. Such a restriction has no bearing on the execution of individual process instances. It becomes relevant only in scenarios of cross-instance process model evolution, where adaptations made within one running instance are intended to propagate to future instances.

\paragraph{Hook cost and FRAME authoring.}
Policy-driven hooks invoke an LLM on every traversal of their attached flow,
introducing per-instance latency proportional to hook count. Two promising mitigations are: (i) caching hook reasoning results keyed on the relevant state variables, so that recurring state patterns avoid repeated LLM calls, and (ii) deploying lightweight pre-filter classifiers that skip hook reasoning when the state clearly satisfies or clearly violates the hook condition without ambiguity. Authoring a complete FRAME also requires care: conflicting policies
across task, decision, and hook documents may produce incoherent behavior. Automated
FRAME consistency checking (analogous to static analysis of classical BPM
models) is an important open problem.

\paragraph{Behavioral diligence of employed agents.}
A further concern is the behavioral diligence of the agents the process harness employs.
The process harness provides each agent with the process knowledge and policy
context needed to govern its reasoning. However, whether the agent's internal
inference actually conforms to its assigned policy, particularly under complex or
ambiguous inputs, lies beyond the process harness's direct control. Establishing such
conformance, for instance through neuro-symbolic inference that provides formal
guarantees of policy adherence, represents an important direction for future
development.

\subsection{Summary}

The formal contribution of this work is a complete framework for uplifting legacy
workflows into Agentic BPM: the Task--Decision--Flow (TDF) model, specifying the
data schema and execution semantics of the process harness, accompanied by its
realization in CUGA FLO.

We have presented CUGA FLO, the first realization of Agentic BPM, and the
Task--Decision--Flow (TDF) process harness model that serves as its conceptual
underpinning. The central insight is that
structured process execution and open-world adaptability are not in fundamental
tension: they require a process-aware, policy-aware agent that oversees
the process (the FlowAgent) and a principled mechanism for runtime structural
intervention: hooks governed by the FRAME. A Model Context Protocol bridge
(\texttt{MCPFlowBridge}) decouples this reasoning authority from a replaceable
workflow execution engine, so that structural conformance is engine-enforced while
adaptation remains policy-bounded.

The TDF model decomposes LLM reasoning across three agent types with distinct
contracts. TaskAgents assemble tools to reason and perform work under task policies.
DecisionAgents evaluate gateway conditions under decision policies, with a
deterministic first pass that resolves routing without an LLM call when the
condition is fully evaluable. The FlowAgent reasons at the meta-level under hook
policies to decide on structural interventions, ranging from simple continues to
runtime modification of the process topology.

The FRAME, the aggregate policy set, provides a concrete accountability substrate: every LLM call in the system is associated with its governing policy document, and any behavioral deviation is diagnosable by inspecting the policy alongside the recorded process knowledge.

The process harness, as introduced in this work, represents a new form of instrumentation for enterprise AI systems: one that combines the structural rigour of deterministic process execution with the open-world reasoning of policy-governed LLM agents. CUGA FLO is its first realization, establishing both the theoretical foundations and a working implementation. The CUGA FLO codebase, including all
policies and the loan approval example, is available at
\url{https://github.com/cuga-project/cuga-agent/tree/cugaflo/src/cuga/backend/cuga_graph/nodes/cuga_flow}.

\bibliographystyle{plain}
\bibliography{references}

\end{document}